\documentclass[10pt,conference]{IEEEtran}

\usepackage{epsfig}
\usepackage{lipsum,graphicx,subcaption}
\usepackage{multicol}
\usepackage{amsmath,amsthm,amssymb,amsfonts}
\usepackage{amssymb}
\usepackage{bm}
\usepackage[noend]{algpseudocode}
\usepackage{algorithmicx,algorithm}
\newtheorem{definition}{Definition}
  
\begin{document}\sloppy
\def\x{{\mathbf x}}
\def\L{{\cal L}}

\pagestyle{empty}

\title{Tensor-Generative Adversarial Network with Two-dimensional Sparse Coding: Application to Real-time Indoor Localization}

\author{Chenxiao Zhu$^{1}$, Lingqing Xu$^2$, Xiao-Yang Liu$^{2,3}$, and Feng Qian$^{1}$\\ 
$^1$The Center for Information Geoscience, University of Electronic Science and Technology of China\\
$^2$Dept. of Electrical Engineering, Columbia University\\
$^3$Dept. of Computer Science and Engineering, Shanghai Jiao Tong University\\
}
\maketitle
\begin{abstract}
\thispagestyle{empty}
 Localization technology is important for the development of indoor location-based services (LBS). Global Positioning System (GPS) becomes invalid in indoor environments due to the non-line-of-sight issue, so it is urgent to develop a real-time high-accuracy localization approach for smartphones. However, accurate localization is challenging due to issues such as real-time response requirements, limited fingerprint samples and mobile device storage. To address these problems, we propose a novel deep learning architecture: Tensor-Generative Adversarial Network (TGAN). 
 
 We first introduce a transform-based 3D tensor to model fingerprint samples. Instead of those passive methods that construct a fingerprint database as a prior, our model applies artificial neural network with deep learning to train network classifiers and then gives out estimations. Then we propose a novel tensor-based super-resolution scheme using the generative adversarial network (GAN) that adopts sparse coding as the generator network and a residual learning network as the discriminator. Further, we analyze the performance of tensor-GAN and implement a trace-based localization experiment, which achieves better performance. Compared to existing methods for smartphones indoor positioning, that are energy-consuming and high demands on devices, TGAN can give out an improved solution in localization accuracy, response time and implementation complexity. 
 
\end{abstract}



\section{Introduction}
\label{sec:intro}
 
 In recent years, the world has witnessed a booming number of smartphones, which apparently has had a profound influence on personal lives. Meanwhile, the application of Internet of Things (IoT) has promoted the development of information technology. The real leap forward comes through the combination of IoT and mobility, which offers us opportunities to re-imagine the ideal life. One promising application of IoT on mobile devices is indoor positioning. More recently, location-based services (LBS) have been used in airports, shopping malls, supermarkets, stadiums, office buildings, and homes \cite{junglas2008location}\cite{wang2014eyes}. Generally, there are three approaches for indoor localization systems, viz., cell-based approach, model-based approach, and fingerprint-based approach. In this paper, we will focus on radio frequency (RF) fingerprint-based localization, which is one of the most promising approaches.

\begin{figure}
    \centering
    \includegraphics[totalheight=6.5cm]{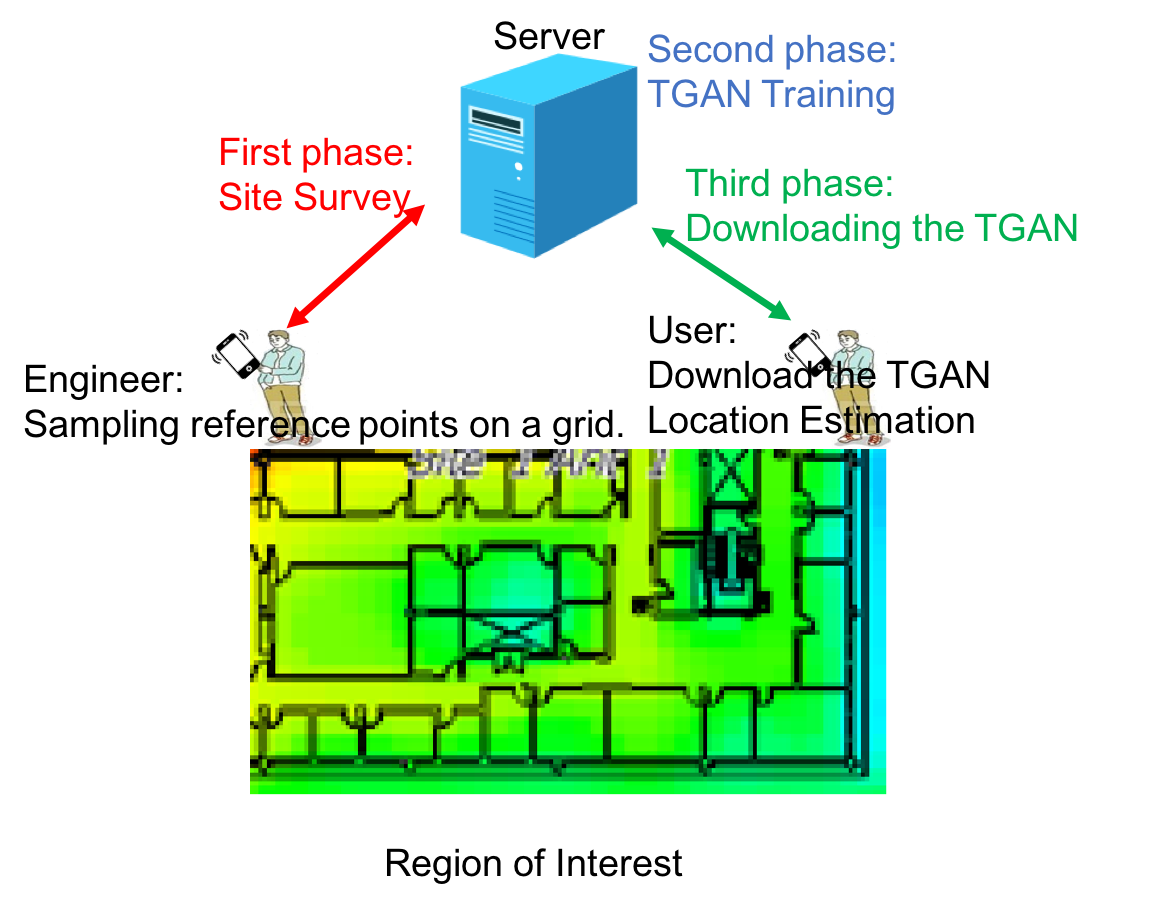}
    \caption{Our indoor localization scenario. The engineer samples RF fingerprints in ROI on a grid, then the server uses samples to train a classification TGAN. The user downloads the trained TGAN and uses a newly sampled fingerprint to estimate current location.}
    \label{fig:indoor localization scenario}
\end{figure}

 RF fingerprint-based indoor localization \cite{liu2016adaptive}\cite{yang2012locating} consists of two phases: training and operating. In the training phase, radio signal strength (RSS) fingerprints are collected from different access points (APs) at reference points (RPs) in the region of interest (ROI). The server uses the collected fingerprint samples to train a TGAN, i.e., taking the fingerprint samples as feature vectors while RPs' coordinates as labels. Then in the operating phase, a user downloads the trained GAN and uses the newly sampled fingerprint to obtain the location estimation. Fig \ref{fig:indoor localization scenario} demonstrates the schema of the indoor localization scenario. 

Existing works on RF fingerprint-based indoor localization have the following limitations. First, for the training phase, constructing a fingerprint database as a prior is unpractical due to lack of high-granularity fingerprint samples we can collect \cite{liu2016adaptive}\cite{yang2012locating}. Secondly, for a mobile device, storage space may not meet the minimum requirement of storing fingerprint database, and the computing capability, usually at tens or hundreds of Million Floating-point Operations per Second (MFLOPs), is also insufficient when encounter such data. Relying on server for storage and computation raises the challenge of communication delay for real-time localization application. In this context, we exploit a novel indoor localization approach based on deep learning. 

Inspired by the powerful performance of generative adversarial networks (GAN) \cite{goodfellow2014generative}\cite{ledig2016photo}, we propose a tensor-based generative adversarial network (TGAN) to achieve super-resolution of the fingerprint tensor. The advantage of GAN lies in recovering the texture of input data, which is better than other super-resolution approaches. We also adopt transform-based approach introduced in \cite{LiuW17b} to construct a new tensor space by defining a new multiplication operation and tensor products. We believe with these multilinear modeling tool, we can process multidimensional-featured data more efficient with our tensor model.

Our paper is organized as follows. Section II presents details about transform-based tensor model with real-world trace verification. Section III describes the architecture of TGAN and gives two algorithms for high-resolution tensor derivation and the training process. Section IV introduces a evaluation of TGAN and implements a trace-based localization experiment. Conclusions are made in Section V.

\section{Transform-Based Tensor Model}

\begin{figure}
    \centering
    \includegraphics[totalheight=6cm]{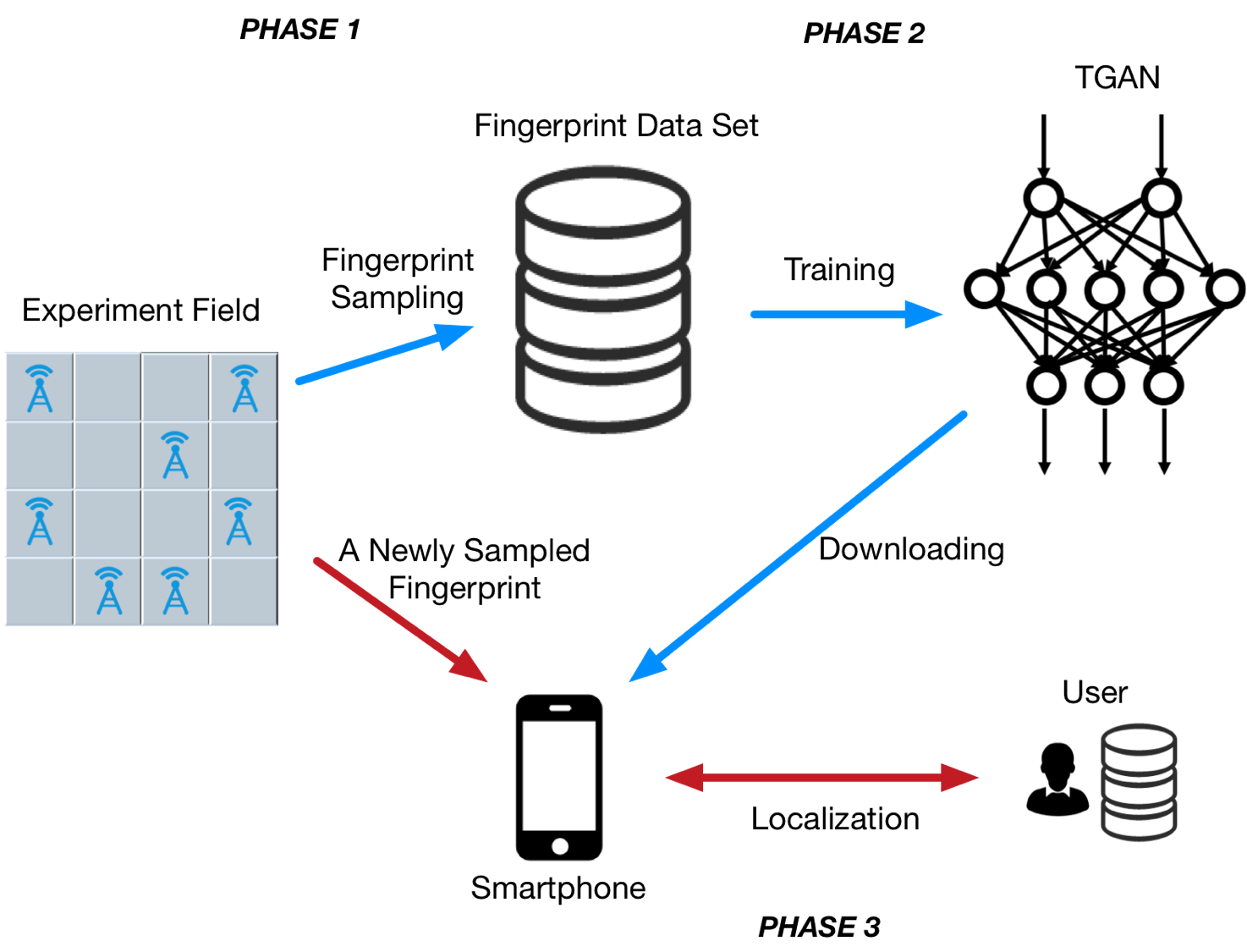}
    \caption{Our localization process can be separated into three phases; phase 1 is the fingerprint sampling as \cite{liu2016adaptive} and phase 2 is the training process of our TGAN. In phase 3, the user downloads the trained TGAN, inputs a newly sampled fingerprint and obtains the location estimation.}
    \label{fig:network}
\end{figure}

\begin{figure}
    \centering
    \includegraphics[totalheight=6cm]{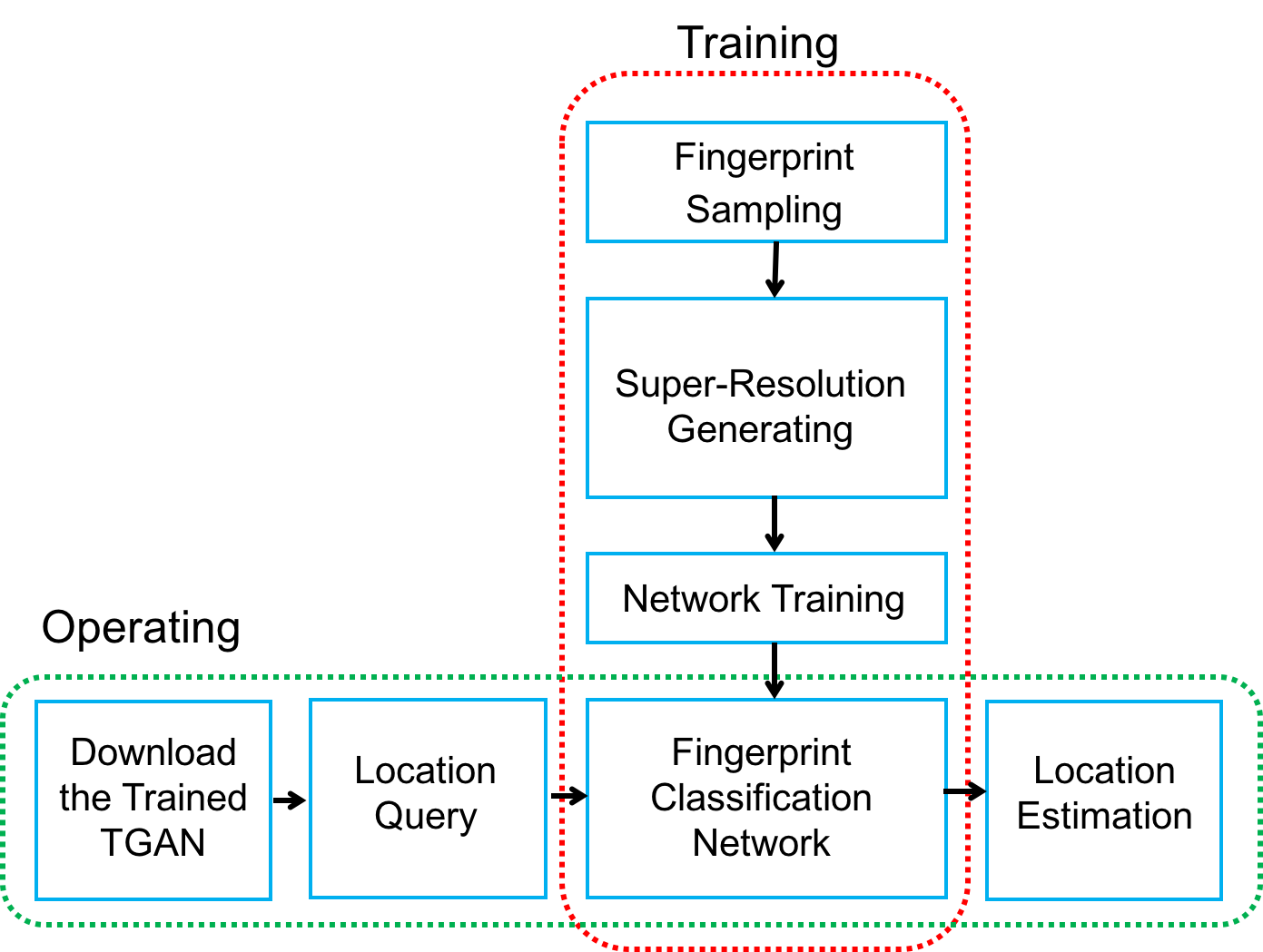}
    \caption{The training phase is completed by engineers and the operation phase by users.}
    \label{fig:network}
\end{figure}

 In this section, we model RF fingerprints as a 3D low-tubal-rank tensor. We begin by first outlining the notations and taking a transform-based approach to define a new multiplication operation and tensor products under multilinear tensor space as in \cite{liu2016adaptive}. We suppose, in the multilinear tensor space, multidimensional-order tensors act as linear operators which is alike to conventional matrix space. In this paper, we use a third-order tensor $\mathcal{T}^{n_1 \times n_2 \times n_3}$ to represent the RSS map. Specifying the discrete transform of interest to be a discrete Fourier transform, the $\bullet$ operation represents circular convolution. 
 
 

 \hspace*{0.02in} {\bf Notations-} In this paper, scalars are denoted by lowercase letters. e.g. $n$; vectors are denoted by boldface lowercase letters e.g. $\bm{a}$ and the transpose is denoted as $\bm{a}^{\top}$; matrices are denoted by boldface capital letters e.g. $\bm{A}$; and tensors are denoted by calligraphic letters e.g. $\mathcal{T}$ and the transpose is denoted by $\mathcal{T}^{\top}$. We use $[n]$ to denote the set $\{1, 2, ..., n\}$. The $\ell_1$ norm of tensor is denoted as $\lVert \mathcal{T}\rVert_1 = \sum_{i,j,k}|\mathcal{T}(i,j,k)|$. The Frobenius norm of a tensor is defined as $\lVert \mathcal{T}\rVert_F = \left( \sum_{i,j,k}\mathcal{T}(i,j,k)^2\right)^{1/2}$ .

\begin{definition}{\bf Tensor product}: For two tensors $\mathcal{A} \in \mathbb{R}^{n_1 \times n_2 \times n_3}$, and $\mathcal{B} \in \mathbb{R}^{n_2 \times n_4 \times n_3}$. The tensor product $\mathcal{C} = \mathcal{A} * \mathcal{B} \in \mathbb{R}^{n_1 \times n_4 \times n_3}$. The $\mathcal{C}(i,j,:)$ is a tube given by $\mathcal{C}(i,j,:) = \sum_{l=1}^{n_2} \mathcal{A}(i,l,:) \bullet \mathcal{B}(l,j,:)$, for $i \in [n_1]$ and $j = [n_4]$.
\end{definition}

\begin{definition}{\bf Identity tensor}: A tensor $\mathcal{I} \in \mathbb{R}^{n_1 \times n_2 \times n_3}$ is identity if $\mathcal{I}(:,:,1)$ is identity matrix of the size $n_1 \times n_1$ and other tensor $\mathcal{I}(:,:,i), i \in [n_3]$ are all zeros.
\end{definition}

\begin{definition}{\bf Orthogonal tensor}: $\mathcal{T} \in \mathbb{R}^{n_1 \times n_2 \times n_3}$ is orthogonal if $\mathcal{T} * \mathcal{T}^{\top} =  \mathcal{T}^{\top} * \mathcal{T} = \mathcal{I}$.
\end{definition}

\begin{definition}{\bf f-diagonal tensor}: A tensor is f-diagonal if all $\mathcal{T}(:,:,i), i \in [n_3]$ are diagonal matrices.
\end{definition}

\begin{definition}{\bf t-SVD}: $\mathcal{T} \in \mathbb{R}^{n_1 \times n_2 \times n_3},$ can be decomposed as $\mathcal{T} = \mathcal{U} * \bm{\Theta} * \mathcal{V}^{\top}$, where $\mathcal{U}$ and $\mathcal{V}$ are orthogonal tensors of sizes $n_1 \times n_1 \times n_3$, $n_2 \times n_2 \times n_3$ individually and $\bm{\Theta}$ is a rectangular f-diagonal tensor of size $n_1 \times n_2 \times n_3$.
\end{definition}

\begin{definition}{\bf Tensor tubal-rank}: The tensor tubal-rank of a third-order tensor is the number of non-zero fibers of $\bm{\Theta}$ in t-SVD.
\end{definition}

\begin{figure}
    \includegraphics[totalheight=4cm]{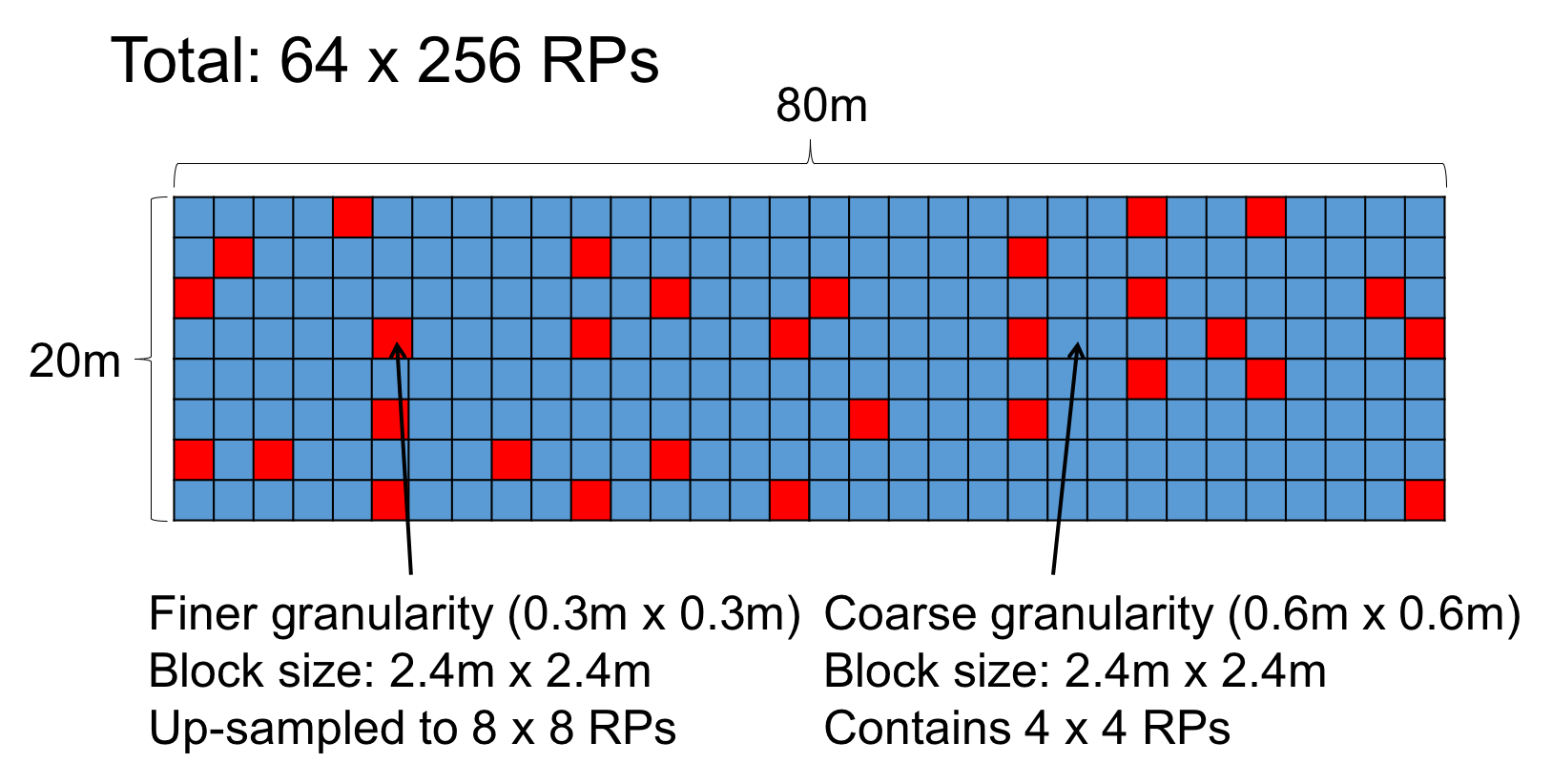}
    \caption{There are $64 \times 256$ RPs uniformly distributed in the $20 \text{m} \times 80\text{m}$ region. The blue blocks represent the finer granularity ones and the red blocks represent the coarse granularity ones}
    \label{fig:scenario}
 \end{figure}

 In this framework, we can apply t-SVD \cite{liu2016adaptive}\cite{LiuW17b}, which indicates that the indoor RF fingerprint samples can be modeled into a low tubal-rank tensor. \cite{LiuW17b} gives out the decomposition measurement of t-SVD regarding normal matrix-SVD, which explains t-SVD may be a suitable approach for further processing. To verify this low-tubal-rank property, we obtain a real-world data set from \cite{liu2016tensor} in an indoor region of size $20\text{m} \times 80\text{m} $ represented as Fig. \ref{fig:scenario}, located in a college building. 
  
\begin{figure}
    \centering
    \includegraphics[totalheight=5cm]{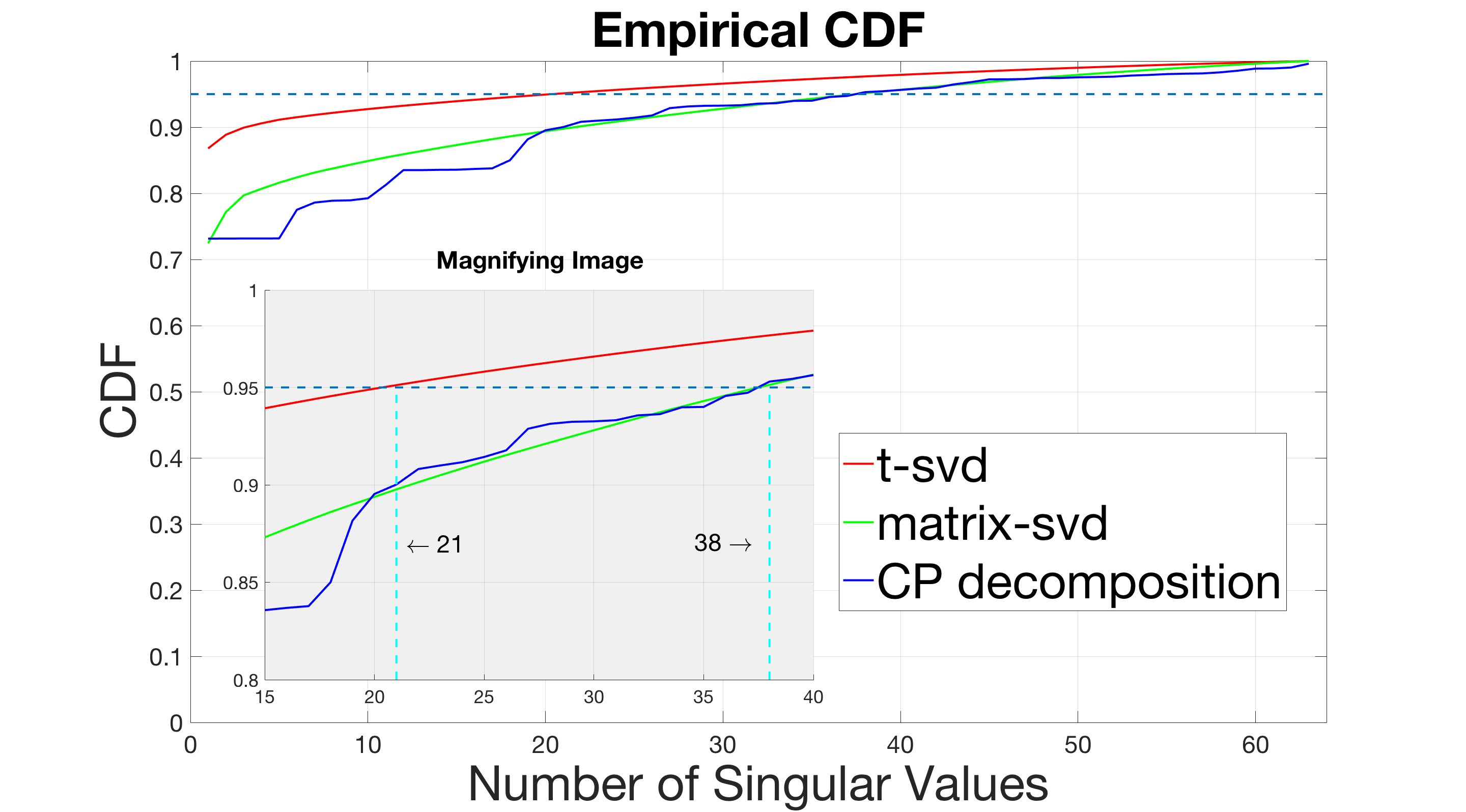}
    \caption{The CDF of singular values for t-SVD, matrix-SVD and CP decomposition. It shows that the low-tubal-rank tensor model better captures the correlations of RF fingerprints.}
    \label{fig:CDF_SVD}
\end{figure}

 The region is divided into $64 \times 268$ grid map with gird size $0.3\text{m} \times 0.3\text{m}$, and there are 21 APs randomly deployed. Therefore, our ground truth tensor is of size $64 \times 256 \times 21$. The Fig. \ref{fig:CDF_SVD} demonstrates that compared with matrix-SVD and CP decomposition \cite{liu2015ls}, the t-SVD shows RF fingerprint data are more likely to be in the form of a low tubal-rank structure. For t-SVD, 21 out of the total 64 singular values capture $95\%$ energy of the fingerprint tensor, and the amount is increased to 38 for the matrix-SVD method and CP decomposition method. Therefore, the low tubal-rank property of transform-based tensor model is more suitable for the RSS fingerprint data than matrix.

 In this section, we compare three decomposition methods and decide to utilize the transform-based tensor model that are constructed under a new tensor space with 2D sparse coding. With the tensor calculations defined \cite{LiuW17b}, we are able to extend the conventional matrix space to third-order tensors. This framework enables a wider tensor application. In the next section, we will introduce generative adversarial networks (GAN) to implement super-resolution based on the tensor model we propose.

\section{Tensor Generative Adversarial Network}

\begin{figure}
    \centering
    \includegraphics[totalheight=7.5cm]{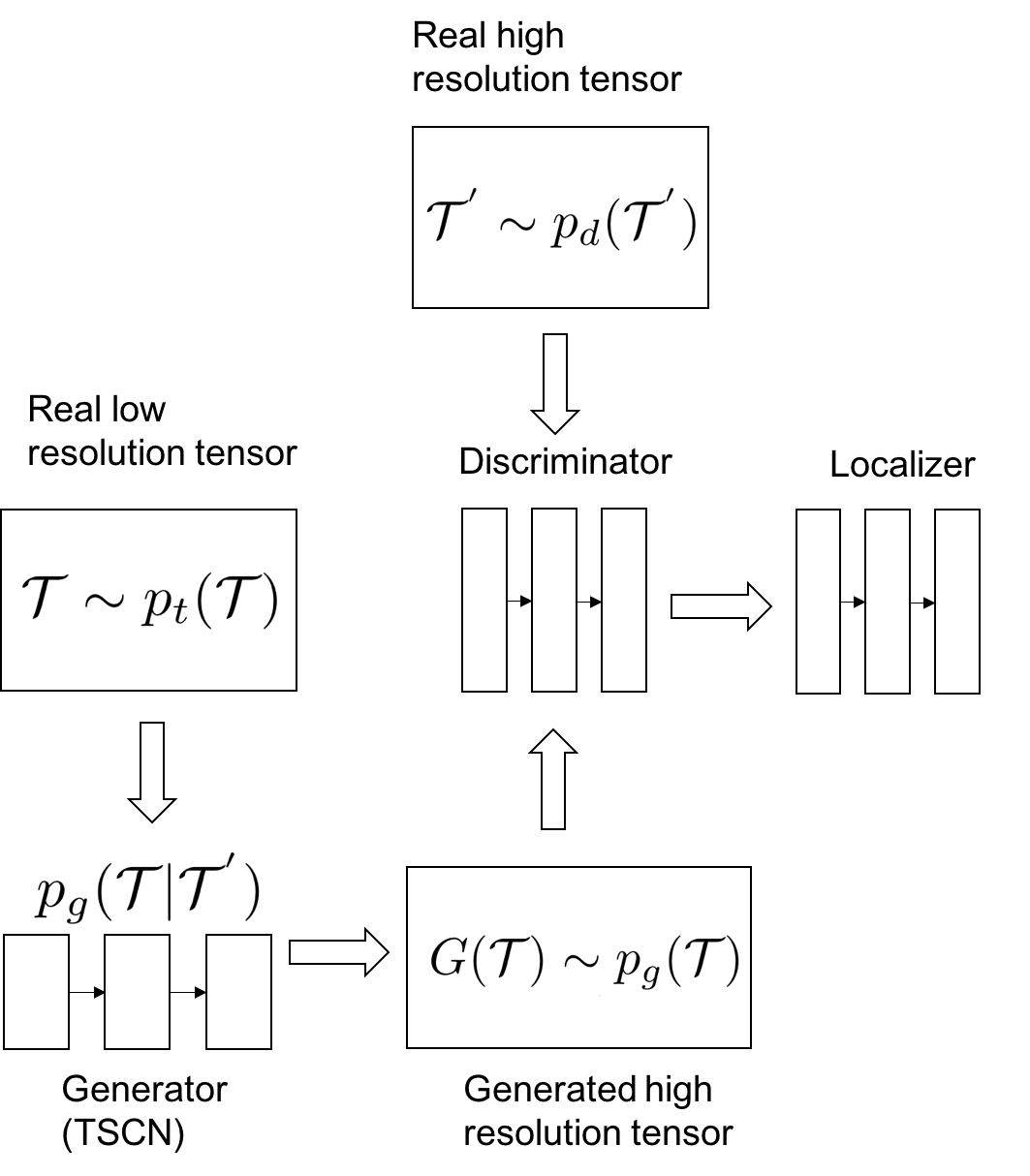}
    \caption{The architecture, data delivery and parameter updating of TGAN.}
    \label{fig:TGAN}
\end{figure}

 TGAN is used to generate extra samples from coarse RF fingerprint samples. The ultimate goal is to train a generator $G$ that estimates a finer granularity tensor $\mathcal{T}^{'}$ from a coarse granularity input tensor $\mathcal{T}$. TGAN consists of three parts: the generator named tensor sparse coding network (TSCN), the discriminator and the localization network named localizer. In this section, we introduce whole architecture of TGAN and illustrates the data delivery and parameter updating process in TGAN as show in Fig. \ref{fig:TGAN}.

 TSCN is used to encode the input coarse granularity tensor $\mathcal{T}$ into a sparse representation $\mathcal{A}$ and its dictionary $\mathcal{D}$ with feedback from discriminator. Therefore, TSCN can be regarded as two parts. The first step is figuring out the sparse representation $\mathcal{A}$ and the dictionary $\mathcal{D}$ by implementing Learned Iterative Shrinkage and Thresholding Algorithm based on Tensor (LISTA-T) and three fully-connection layers of network to receive feedback from discriminator. Then the output of the LISTA-T algorithm is modified. This operation scheme is similar to the function of sparse autoencoder, since the TSCN's target function (\ref{equation:Lagrange})  has the same format with autoencoder.

 Let $\mathcal{T^{'}}$ be the input, $p_g(\mathcal{T^{'}})$ be the model distribution and  $p_t(\mathcal{T^{'}})$ be the data distribution. $p_g(\mathcal{T}|\mathcal{T^{'}})$ can be regard as encoding distribution of auto encoder, or as TSCN correspondingly. Therefore, we can define that $p_g(\mathcal{T^{'}}) = \int_{\mathcal{T}} p_g(\mathcal{T}|\mathcal{T^{'}}) ~ p_t(\mathcal{T}) d\mathcal{T}$. Let $p_d(\mathcal{T})$ be the arbitrary prior distribution of $\mathcal{T}$. By matching the aggregated posterior $p_g(\mathcal{T})$ to an arbitrary prior distribution $p_d(\mathcal{T}^{'})$, the TSCN can be regularized. In the following work, TSCN parametrized by $\theta_G$ is denoted as $G_{\theta_G}$. Then, finding a suitable TSCN for generating a finer granularity tensor ${T}^{'}$ out of coarse tensor ${T}$ can be represent as: $\widehat{\theta}_G = \text{arg min}~ \frac{1}{L} \sum^{L}_{l=1} \lVert G_{\theta_G}(\mathcal{T}_l),\mathcal{T}^{'}_l \rVert^{2}_{F}$. The target adversarial min-max problem is:

\begin{equation}
\begin{aligned}
\label{equation: GAN target}
\text{min}_{\theta_G}\text{max}_{\theta_D} & \mathbb{E}_{\mathcal{T}^{'} p_d(\mathcal{T}^{'})} \log{ D_{\theta_d}(\mathcal{T}^{'})}  +\\
&  \mathbb{E}_{\mathcal{T} p_g(\mathcal{T})} \log{( 1 -  D_{\theta_d}(G_{\theta_G}(\mathcal{T})))},
\end{aligned}
\end{equation}

 The encoding procedure from a coarse granularity tensor $\mathcal{T}$ to a higher resolution tensor $\mathcal{T}^{'}$ of a tensor can be regarded as a super-resolution problem. The problem is ill-posed and we model two constraints to solve it. The first one is that the finer granularity patches can be sparsely represented in an appropriate dictionary and that their sparse representations can be recovered from the coarse granularity observation. The other one is the recovered tensor $\mathcal{T}^{'}$ must be consistent with the input tensor $\mathcal{T}$ for reconstruction. Corresponding to $\mathcal{T}$ and $\mathcal{T}^{'}$, dictionaries are defined $\mathcal{D}$ and $\mathcal{D}^{'}$. In our algorithm, the approach to recover $\mathcal{T}^{'}$ from $\mathcal{T}$ is to keep $\mathcal{D}^{'}$ and $\mathcal{D}$ as a same sparse representation $\mathcal{A}$. To simplify the tensor model, we should minimize the total number of coefficients of sparse representation, which can be represent as $\min ~\lVert \mathcal{A} \rVert_0,~\text{such that}~ \mathcal{T}^{'} \approx \mathcal{D} * \mathcal{A}$, where higher resolution tensor $\mathcal{T}^{'} \in \mathbb{R}^{n_1 \times n_2 \times n_3}$, sparse representation coefficient $\mathcal{A} \in \mathbb{R}^{n_1 \times n_2 \times n_3}$ and dictionary $\mathcal{D} \in \mathbb{R}^{n_1 \times n_2 \times n_3}$.

 Suggested by \cite{donoho2006most}, in order to make $\mathcal{A}$ sparse, we present the optimization equation: equation (\ref{equation:Lagrange}). It can be simplified by Lagrange multiplier $\lambda$ to equation  (\ref{equation:Lagrange1}). Here, $\lambda$ balances sparsity of the solution and fidelity of the approximation to $\mathcal{T}$.
 
\begin{equation}
\min\limits~ \lVert \mathcal{A}\rVert_1 ~~ \text{s.t.} ~ \lVert \mathcal{D} *  \mathcal{A} - \mathcal{T}\rVert^{2}_F \le \epsilon.
\label{equation:Lagrange}
\end{equation}

\begin{equation}
\min\limits_{\mathcal{A}} ~\lVert \mathcal{D} *  \mathcal{A} - \mathcal{T}\rVert^{2}_F + \lambda \lVert \mathcal{A} \rVert _1 ,
\label{equation:Lagrange1}
\end{equation}

Recent researches \cite{wang2015deep}\cite{kavukcuoglu2010fast}\cite{gregor2010learning} indicate that there is an intimate connection between sparse coding and neural network. In order to fit in our tensor situation, we propose a back-propagation algorithm called learned iterative shrinkage and thresholding algorithm based on tensor (LISTA-T) to efficiently approximate the sparse representation (sparse coefficient matrix) $\mathcal{A}$ of input $\mathcal{T}$ via using trained dictionary $\mathcal{D}$ to solve equation (\ref{equation:Lagrange}). The entire super-resolution algorithm described above is illustrated in Algorithm 1.

\begin{algorithm}[t]
\caption{Super-resolution via Sparse Coding}
\hspace*{0.02in} {\bf Input:}
trained dictionaries $\mathcal{D}^{'}$ and $\mathcal{D}$, a coarse granularity fingerprint image $\mathcal{Y}$.\\
\hspace*{0.02in} {\bf Output:} 
Super-Resolution fingerprint image $\mathcal{X}$.
\begin{algorithmic}[1]
\State Solve $\mathcal{D}$ and $\mathcal{D}^{'}$ via algorithm 2.
\For{Each patch of $\mathcal{Y}$}
　　\State Use $\mathcal{D}$ to solve the optimization problem for $\mathcal{A}$ in (\ref{equation:Lagrange}),
　　\State Generate the finer granularity patch $\mathcal{X} = \mathcal{D}^{'} * \mathcal{A}$.
\EndFor
\end{algorithmic}
\end{algorithm}

 Accompanied with the resolution recovering procedure to obtain pairs of high and coarse granularity tensor patches that have the same sparse representations, we should solve the respected two dictionaries $\mathcal{D}^{'}$ and $\mathcal{D}$. We present an efficient way to learn a compact dictionary by training single dictionary.

 Sparse coding is a problem of finding sparse representations of the signals with respect to an overcomplete dictionary $\mathcal{D}$. The dictionary is usually learned from a set of training examples $\mathcal{T}^{'} = \{ \mathcal{T}^{'}_1,\mathcal{T}^{'}_2,...,\mathcal{T}^{'}_l\}$. To optimize coefficient matrix $\mathcal{A}$, we first fix $\mathcal{D}$ and utilize ISTA-T to solve the optimization problem. That is: 

\begin{equation}
\label{equation:optimize B}
\min\limits_{\mathcal{A}} f(\mathcal{A}) + \lambda g(\mathcal{A}) ,
\end{equation}

 where $f(\mathcal{A})$ stands for the data reconstruction term $\lVert \mathcal{T}^{'} - \mathcal{D}^{'} * \mathcal{A}^{'} \rVert^2_F$, the Forbenius norm constraints on the term remove the scaling ambiguity, and $g(\mathcal{A})$ stands for the sparsity constraint term $\lVert \mathcal{A}^{'} \rVert_1$. 

 ISTA-T is used to solve this problem, which can be rewritten as a linearized, iterative functional around the previous estimation $\mathcal{A}_p$ with proximal regularization and non-smooth regularization. Thus at the $(p+1)$-th iteration, $\mathcal{A}_{p+1}$ can be update by equation (\ref{equation:Ap1}). By using $L_{p+1}$ to be Lipschitz constant, and $\nabla f(\mathcal{A})$ to be gradient defined in the tensor space, we can have equation (\ref{equation:Ap1L})

\begin{algorithm}[t]
\caption{Dictionary Training based on Tensor} 
\hspace*{0.02in} {\bf Input:} 
finer granularity tensor $\mathcal{T}^{'}$ and the maximum iterations: num.\\ 
\hspace*{0.02in} {\bf Output:} 
Trained dictionary $\mathcal{D}^{'}$.
\begin{algorithmic}[1]
\State Initialize $\mathcal{D}^{'}$ with a Gaussian random tensor,
\For{$t = 1$ to num}
    \State Fix $\mathcal{D}^{'}$ and solve $\mathcal{A}$ via (\ref{equation:optimize B}) to (\ref{equation: optimize B final}),
    \State Fix $\mathcal{A}$ and set $\widehat{\mathcal{T}^{'}} = \text{fft}(\mathcal{T}^{'},[~],3), \widehat{\mathcal{A}} = \text{fft}(\mathcal{A},[~],3)$,
\EndFor
\For{$k = 1$ to $n_3$}
    \State Solving (22) for $\Lambda$ by Newton's method,
    \State Calculate $\widehat{\mathcal{D}}^{'}$ from (\ref{equation:LDAA}),
    \State $\mathcal{D}^{'} = \text{ifft}(\widehat{\mathcal{D}}^{'},[~],3)$.
\EndFor
\end{algorithmic}
\end{algorithm}

\begin{equation}
\begin{aligned}
\label{equation:Ap1}
\mathcal{A}_{p+1} = \text{arg} \min\limits_{\mathcal{A}} f(\mathcal{A}_p) + \langle \nabla f(\mathcal{A}_p), \mathcal{A} - \mathcal{A}_p \rangle \\
+ \frac{L_{p+1}}{2} \lVert \mathcal{A} - \mathcal{A}_p \rVert^2_F + \lambda g(\mathcal{A})  ,
\end{aligned}
\end{equation}

\begin{equation}
\begin{aligned}
\label{equation:Ap1L}
\mathcal{A}_{p+1} = \text{arg} \min\limits_{\mathcal{A}} \frac{1}{2} \lVert \mathcal{A} - (\mathcal{A}_p -\frac{1}{L_{p+1}} \nabla f(\mathcal{A}_p)) \rVert^2_F \\+ \frac{\lambda}{L_{p+1}}\lVert \mathcal{A}\rVert_1  .
\end{aligned}
\end{equation}
  
 Here, $\nabla f(\mathcal{A}) = \mathcal{D}^{\top} * \mathcal{D} * \mathcal{A} - \mathcal{D}^{\top} - \mathcal{T}^{'}$. As we introduced in section II, the transform-based tensor model can also be analyzed in the frequency domain, where the tensor product $\mathcal{T} = \mathcal{D} * \mathcal{B}$. It can be computed as $\widehat{\mathcal{T}}^{(k)} = \widehat{\mathcal{D}}^{(k)} \widehat{\mathcal{B}}^{(k)}$, where $\mathcal{T}^{(k)}$ denotes the k-th element of the expansion of $\mathcal{X}$ in the third dimension and $k \in [n_3]$. For every $\mathcal{A}_{p+1}$ and $\mathcal{A}_{p}$, we can conduct that:
\begin{equation}
\label{equation: optimize B final}
\begin{aligned}
&\lVert \nabla f(\mathcal{A}_{p+1}) - \nabla f(\mathcal{A}_p) \rVert_F \\
\le \sum^{n_3}_{k=1} & \lVert \widehat{\mathcal{D}}^{(k)^{\top}} \widehat{\mathcal{D}}^{(k)} \rVert^2_F\lVert\mathcal{A}_{p+1}-\mathcal{A}_p\rVert_F .
\end{aligned}
\end{equation}

Thus Lipschitz constant for $f(\mathcal{A})$ in our algorithm is:
\begin{equation}
L_{p+1} = \sum^{n_3}_{k=1}\lVert \widehat{\mathcal{D}}^{(k)^{\top}} \widehat{\mathcal{D}}^{(k)} \rVert^2_F
\end{equation}

\begin{figure*}[t]
\begin{multicols}{3}
    \includegraphics[width=6cm,height=3.3cm]{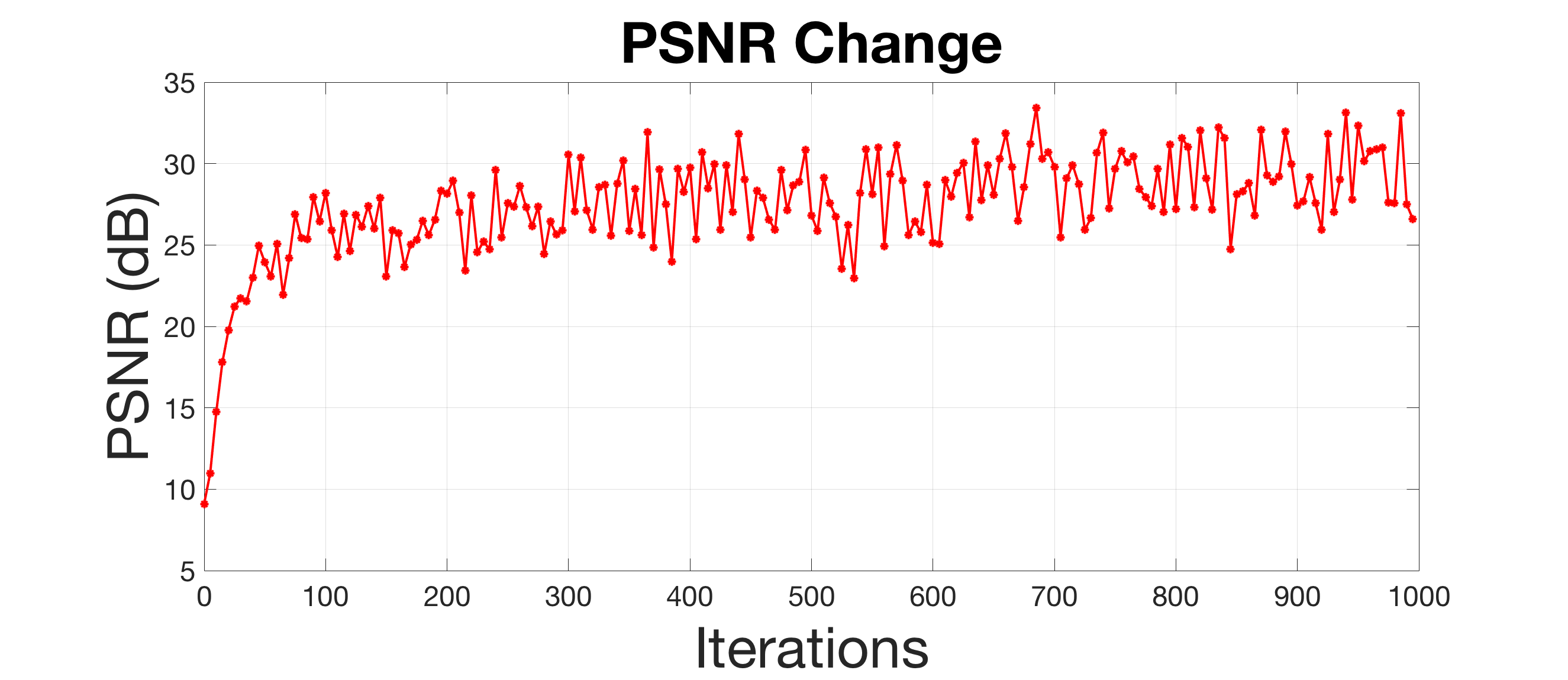}\par\caption{The PSNR change in training procedure.}\label{fig:PSNR}
    \includegraphics[width=6.5cm,height=3.3cm]{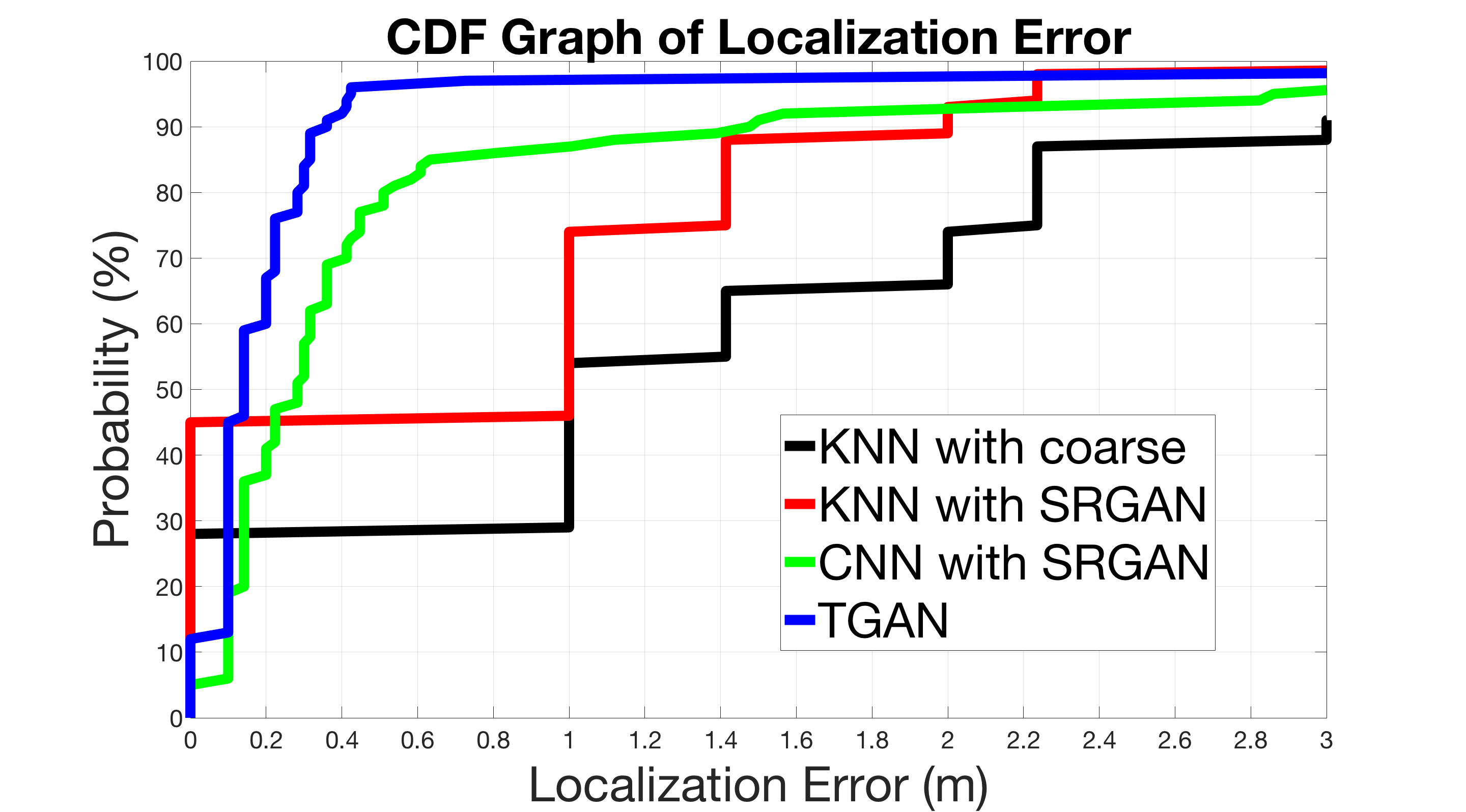}\par\caption{CDF of location estimation error}\label{fig:CDF_Final}
    \includegraphics[width=6cm,height=3.3cm]{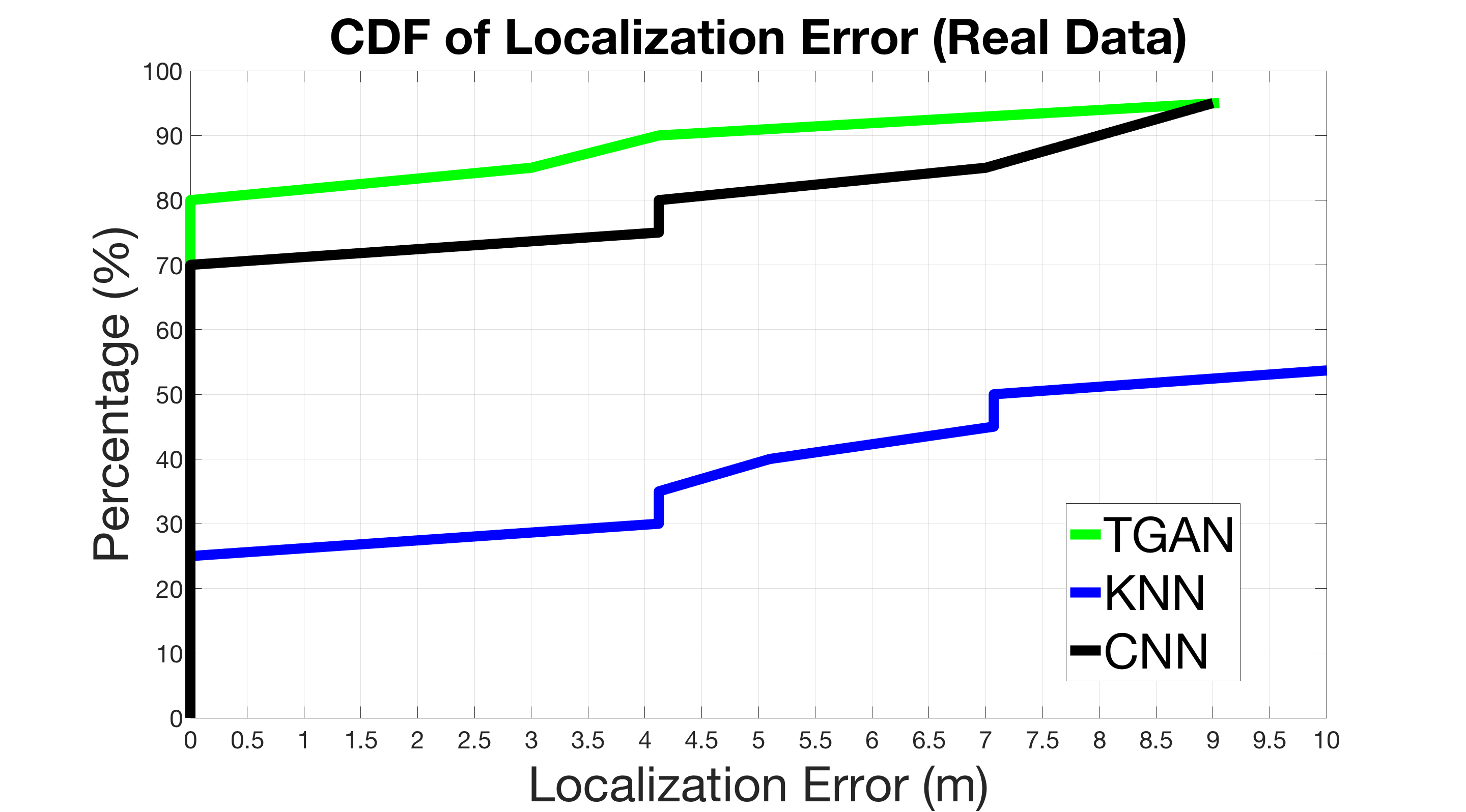}\par\caption{CDF of localization error of TGAN}\label{fig:real_data}
\end{multicols}
\end{figure*}



Then we can optimized $\mathcal{A}$ by proximal operator $\text{Prox}_{\lambda/ L_{p+1}} (\mathcal{A}_p - \frac{1}{L_{p+1}} \nabla f(\mathcal{A}_p)) $, where $\text{Prox}_\tau(\cdot) = \text{sign}(\cdot) \text{max} (| \cdot | - \tau,0) $ is the soft-thresholding operator \cite{liu2016tensor}. After that, we should fix $\mathcal{A}$ to update corresponding dictionary $\mathcal{D}$. We rewrite equation (\ref{equation:optimize B}) as equation (\ref{equation:Dh defination}). For the reason that, in this paper, we specify the discrete transform of interest to be a discrete Fourier transform, aka., decomposing it into k-nearly-independent problems using DFT like equation (\ref{equation:DFT}). Here, $\widehat{\mathcal{T}^{'}}$ denotes the frequency representation of $\mathcal{T}$, i.e. $\widehat{\mathcal{T}^{'}} = \text{fft}(\mathcal{T}, [~], 3)$.

\begin{equation}
\begin{aligned}
    \mathcal{D}^{'},\mathcal{A}^{'} = ~ &\text{arg} \min\limits_{\mathcal{D}^{'},\mathcal{A}^{'}} \lVert \mathcal{T}^{'} - \mathcal{D}^{'} * \mathcal{A}^{'} \rVert^2_F + \lambda \lVert \mathcal{A}^{'} \rVert_1 \\
    &\text{s.t.} ~\lVert \mathcal{D}_i (:, j , :) \rVert^2_F \le 1, j \in [r],
\end{aligned}
\label{equation:Dh defination}
\end{equation}

\begin{equation}
\begin{aligned}
&\min\limits_{\widehat{\mathcal{D}}} ~~ \sum^{n_3}_{k=1} \lVert \widehat{\mathcal{T}^{'}}^{(k)} - \widehat{\mathcal{D}}^{(k)}\widehat{\mathcal{A}}^{(k)} \rVert^2_F \\
&\text{s.t.}~ \sum^{n_3}_{k=1}\lVert \widehat{\mathcal{D}}^{(k)}(:,j)\rVert^2_F \le r, j \in [r] ,
\end{aligned}
\label{equation:DFT}
\end{equation}

We can apply Lagrange dual with a dual variable $\lambda_j \ge 0, j \in [r]$ to solve equation (\ref{equation:DFT}) in frequency domain as equation (\ref{equation:LDA}). By defining $\Lambda = \text{diag}(\lambda)$, the minimization over $\widehat{\mathcal{D}}$ can be write as equation (\ref{equation:LDAA}). Substituting the equation (\ref{equation:LDAA}) into Lagrangian $\mathcal{L}(\widehat{\mathcal{D}},\Lambda)$, we will obtain the Lagrange dual function $\mathcal{L}(\Lambda)$ as equation (\ref{equation:LDAAA}). Notation $\text{Tr}(\cdot)$ represents the trace norm of a matrix.

\begin{equation}
\begin{aligned}
\mathcal{L}(\widehat{\mathcal{D}}, \Lambda) &= \sum^{n_3}_{k=1} \lVert \widehat{\mathcal{T}^{'}}^{(k)} - \widehat{\mathcal{D}}^{(k)}\widehat{\mathcal{A}}^{(k)} \rVert^2_F \\ 
&+ \sum^{r}_{j=1} \lambda_j \left(\lVert\widehat{D}(:,j,:)\rVert_F^2 -r\right),
\label{equation:LDA}
\end{aligned}
\end{equation}

\begin{equation}
\widehat{\mathcal{D}}^{(k)} = (\widehat{\mathcal{T^{'}}^{(k)}}\widehat{\mathcal{A}}^{(k)^{\top}})(\widehat{\mathcal{A}}^{(k)}\widehat{\mathcal{A}}^{(k)^{\top}} + \Lambda)^{-1}, k \in [n_3] .
\label{equation:LDAA}
\end{equation}

\begin{equation}
\mathcal{L}(\Lambda) = -\sum^{n_3}_{k=1}\text{Tr}(\widehat{\mathcal{A}}^{(k)^{\top}}\widehat{\mathcal{T}^{'}}^{(k)}\widehat{\mathcal{D}}^{(k)^{\top}}) -k \sum^{r}_{j=1}\lambda_j ,
\label{equation:LDAAA}
\end{equation}

 Equation (\ref{equation:LDAAA}) can be solved by Newton's method or conjugate gradient. After we get the dual variables $\Lambda$, the dictionary can be recovered. The whole training algorithm of dictionary $\mathcal{D}$ is summarized in Algorithm 2. 
 
 With the input finer granularity tensor $\mathcal{T}^{'}$, the algorithm can return the trained dictionary $\mathcal{D}^{'}$ and with the input coarse granularity tensor $\mathcal{T}$, the algorithm returns $\mathcal{D}$. So, this is an efficient way to learn a compact dictionary by training single dictionary.
 
\section{Performance Evaluation}

In this section, we first compare the super-resolution tensor results that obtained by applying TGAN on trace- based fingerprint tensor data with other algorithms. Then we evaluate the localization accuracy and compare the performance of our localization model with other previous conventional models using the same data. Finally we implement our own experiment and collect another ground-truth fingerprint data set to verify our approach.

As illustrated in Fig. \ref{fig:CDF_SVD}, the trace data set is collected in the region of size $20\text{m} \times 80\text{m}$ and the tensor set can be measured as $64 \times 256 \times 21$. Each block have $4 \times 4$ RPs, we then set some blocks which have $4 \times 4$ RPs of coarse granularity tensor $\mathcal{T}^{'}$ contrasted with the rest which have $8 \times 8$ RPs of finer granularity tensor $\mathcal{T}$ by $4 \times$ up-sampling (The finer granularity of blocks is $0.3\text{m} \times 0.3 \text{m}$). $70\%$ blocks enjoy finer granularity, and $30\%$ blocks ready to be processed by TGAN. For this scenario, to measure the quality of generated finer granularity tensor, we adopt the peak signal-to-noise ratio (PSNR) criteria denoted as:
\begin{equation}
\text{PSNR} = 20\times \log_{10}{\frac{\text{MAX}_I}{\text{MSE}}},
\end{equation}
where $\text{MSE} = \frac{1}{n_1 n_2 n_3} \Vert \mathcal{T}^{'} - \mathcal{T} \rVert_F$

There are several existing super-resolution algorithms with our LISTA-T, as suggested in \cite{ledig2016photo}\cite{shi2016deconvolution}\cite{wang2015deep}, Trilinear interpolation, Super-resolution GAN (SRGAN) and the Sparse coding based networks (SCN) we adopt. The measurements are shown in Fig. \ref{fig:PSNR} and Fig. \ref{fig:CDF_Final}. 

From Fig. \ref{fig:PSNR}, we can tell that the calculation of TGAN converge fast. And we measure the Euclidean distance between the estimated location and the actual location of the testing point as localization error by equation (\ref{equation:dist}). 

Fig. \ref{fig:CDF_Final} describes the cumulative distribution function (CDF) of location estimation error of fingerprint classify network with and without original training samples and KNN algorithm.
\begin{equation}
d_e = \sqrt{(\widehat{x}-x)^2 + (\widehat{y}-y)^2}
\label{equation:dist}
\end{equation}

For trace data in localization performance, in Fig. \ref{fig:CDF_Final}, we adopt KNN and SRGAN to compare with the fingerprint-based indoor localization and draw a CDF graph of localization error. For the location estimation with fingerprint data, both neural networks (TGAN and CNN) perform better than KNN with small variation. The error of localization with KNN will increase stage by stage, while the error in neural network increases smoothly. By comparing with the TGAN and CNN supplied by the SRGAN, the improvement shows that the TGAN performs better than the simple merge of generated extra tensor samples and CNN. The TGAN can catch the interconnection inside the data better and generate more similar data for localization.


In our own experiment, we develop an Android application to collect the indoor Wi-Fi RF fingerprint data in a smaller region. With the native Android WifiManager API introduced in Android developer webpage, we can manage all aspects of Wi-Fi connectivity. To be specific, we select a $6\text{m} \times 16\text{m}$ region with 14 APs lie inside, and we set the scale of the RPs as $1\text{m} \times 1\text{m}$, so RSS fingerprint tensor set is of size $6 \times 16 \times 14$. The raw fingerprint data we collect is 4.2MB.

After collecting data, we choose two representative localization techniques to compare with TGAN, namely, weighted KNN and conventional CNN. Weighted KNN is the most widely used technique and is reported to have good performance for indoor localization systems. The conventional CNN in this experiment is similar to the localization part of TGAN, while TGAN holds an extra step for processing collected coarse data to high granularity ones, which is called Super-resolution via Sparse Coding. The trained TGAN downloaded to smartphone is only 26KB in our experiment. The result is shown in Fig.\ref{fig:real_data}. By comparing the localization error of KNN algorithm, CNN and TGAN, the terrific result shows that the neural network is much suitable for indoor RSS fingerprint data than KNN, which is consistant with the result of our simulation.

Based on the results of simulation and experiment we made above, we can summarize the improvement of indoor localization in our work. First, the transform-based tensor model can handle insufficient fingerprint samples well by sparse representation, which means we can obtain a finer granularity fingerprint with sparse coding. Secondly, we adopt CNN like localization step and it is proved to be much suitable for analyzing the indoor RSS fingerprint data than KNN, which enables TGAN to give out an improved solution in localization accuracy. Thirdly, TGAN is easy to use and proved to reduce the storage cost  of smartphones by downloading the TGAN rather than raw data, so it is acceptable to implement TGAN on smartphones.




\section{Conclusion}

We first adopted a transform-based tensor model for RF fingerprint and, based on this model, we raised a novel architecture for real-time indoor localization by combining the architecture of sparse coding algorithm called LISTA-T and neural network implementation called TSCN. In order to solve the insufficiency of training samples, we introduced the TGAN to complete super-resolution for generating new fingerprint samples from collected coarse samples. In future works, we will try to figure out a more efficient and slim network architecture for localization process. We believe that applying neural network on mobile platforms is promising.

\bibliographystyle{IEEEbib}
\bibliography{reference.bib}
\end{document}